\documentstyle[fullname]{article}    

\title{Some Bibliographical References on\\Intonation and Intonational Meaning}

\author{Julia Hirschberg}

\begin{document}           

\maketitle

\section{Selected Work on Intonational Meaning}
Introductions to speech research and acoustic phonetics can be found
in \cite{Ladefoged62,Denes73,Fry79}. For survey-level introductions to
studies of English intonation, see
\cite{Halliday67b,Crystal69,Crystal75,Cruttenden86,Couper-Kuhlen86}.
\cite{Ladd80,Bolinger86,Bolinger89} also represent good introductions
to studies of intonational meaning.  Works oriented toward particular
aspects of intonation theory and models of intonation, also primarily
for English are
\cite{Bolinger51,Liberman75,Garding83,Hirst83,Pierrehumbert80,Vaissiere83,Liberman84,Ladd88,tHart90}
\nocite{Ladd83a,Ladd83b,Ladd83c,Ladd86,Pike45,O'Connor61}
\nocite{Buxton83,Edwards88}
An idea of some of the topics and methods of research on intonational
representation and intonational meaning can
be gained from \cite{Cutler83a,Gibbon84}.
\nocite{Duncan77}

The meaning of particular intonational contours has been studied by
\cite{Liberman74,Sag75,Ladd77,Ladd78,Bing79b,Ladd80,Bouton82,Ward85b}.
Attempts to determine the relative contribution of intonational
contours and other intonational features to utterance interpretation
include \cite{Ladd85,Pierrehumbert90,Hirschberg91b}.

For intonational studies of adult intonation in speech directed to
children, see the overview in
\cite{Fernald91b}.  For experimental work on infants' perception of
phrasing see \cite{Hirsh-Pasek85,Ratner86}, and for children's
turn-taking devices, see
\cite{Ervin-Tripp79}.  \cite{Morgan87} studies prosodic and
morphological cues to language acquisition.
\nocite{Fernald91a}

There is a considerable literature on the relationship between
intonational phrasing and syntactic phenomena.  General and
theoretical work includes \cite{Downing70,Bresnan71,Selkirk84a,Cooper80}.
Empirical work on acoustic correlates of intonation boundaries
includes experimental work (production and perception) such as
\cite{Grosjean79,O'Malley73,Lehiste73,Klatt75a,Lehiste76,Cooper77,Streeter78,Wales79,Lehiste80a,Gee83,Umeda82,Beach91}.
\nocite{Martin70,Grosjean87}
Corpus-based research includes \cite{Quirk64,Altenberg87}.  Work in
computational linguistics proposes parsing strategies including
prosodic components \cite{Marcus85,Marcus90,Steedman90}.
Text-to-speech applications inspire some of the work on intonational
boundary predictions \cite{Gee83,Altenberg87,Bachenko90}.  Recognition
applications have motivated recent work
\cite{Ostendorf90,Wang92}.
Also see \cite{Wilkenfeld81} for research on prosody and orthography.
\nocite{Silva-Corvalan83,Bruce90}
\nocite{Taft84,O'Shaughnessy89}
\nocite{Wang91a}
Also see \nocite{Misono90,Sugito90}

Early debates on intonational prominence (stress or pitch accent) are
summarized in
\cite{Bolinger58,Crystal69,Liberman77,Ladd80,Bolinger86}.  More recent
contributions include \cite{Beckman86c,Pierrehumbert87b}.  Constraints
on sentence (nuclear) stress are discussed in
\cite{Cutler77b,Erteschik83,Schmerling76,Schmerling74,Bardovi83a}.
Despite Bolinger's seminal article on the unpredictability of accent
\cite{Bolinger72a}, attempts to predict accent from other features of
the uttered text include \cite{Altenberg87,Hirschberg90}.  A number of
authors have examined the relationship between accent and various
characterizations of information status: Work on the focal domains of
accent and the representation and interpretation of intonational focus
and presupposition includes
\cite{Lakoff71,Schmerling71,Jackendoff72,Ball77,Wilson79,Enkvist79,Gussenhoven83a,Culicover83,Rooth85,Rochemont90,Rooth91,Horne85,Horne87,Baart87,Dirksen92,Zacharski92}.
Topic/comment, given/new, theme/rheme distinctions are discussed with
respect to accent by
\cite{Schmerling75b,Bardovi83b,GBrown83a,Gundel78,Lehman77,Fuchs80,Chafe76,Nooteboom82,Fuchs84,Terken84,Terken85,Terken87,Fowler87,Horne91a,Horne91b}.
\nocite{Allerton79,Kruyt85}
And contrastive stress is examined by
\cite{Bolinger61,Harries-Delisle78,Couper-Kuhlen84}.
Others have looked at the interpretation of accent with particular
attention to anaphora
\cite{Gleitman61,Akmajian70,Williams80,Lujan85,Hirschberg91a}.
See \cite{Ladd87a,Sproat90} for discussions of the phrasing and
accenting of complex nominals.
\nocite{Cantrall69,Cantrall75}
\nocite{Hirschberg90a}
\nocite{Wilson79}
\nocite{Pierrehumbert89}
\nocite{Liberman77}

For general discussion of the intonational characteristics of longer
discourses,
see \cite{Brazil80,GBrown80,GBrown83b}.  Halliday \cite{Halliday76}
also has insightful comments on intonation and discourse cohesion.
For work on intonation and discourse structure, see
\cite{Yule80,Silverman86,Avesani88,Ayers92,Hirschberg92b,Grosz92}, and see
\cite{Kumpf87} for discussion of
pitch phenomena and stories.  \cite{Hirschberg87a,Litman90} investigate
the intonation of cue phrases.
\cite{Butterworth75,Butterworth77,Schegloff79b} investigate
intonational characteristics of speech-only communication channels.

\nocite{Goodwin81}
\nocite{Kutik83}
\nocite{Rees76}
\nocite{Hirschberg86} \nocite{Gazdar80b}
\nocite{Gumperz77}

A good overview of work in speech synthesis is \cite{Klatt87}.
Current work in the area can be sampled in \cite{Autrans90}.
\nocite{Olive85,Anderson84,JHouse90a}
Some experimental message-to-speech systems are described in
\cite{Young79,LWitten77,Danlos86,Davis88,JHouse90b}.
A good introduction to past and recent work on speech recognition can
be found in \cite{Waibel90a}.  More specialized work on the use of
prosody in recognition includes
\cite{Lea75,Lea79,Pierrehumbert83,Waibel79,Ljolje86,Ljolje87a,Waibel88,DHouse89,Silverman92a}
\nocite{Pierrehumbert81,Hirschberg89}
\nocite{Silverman87}

\nocite{Hemphill90,Price88}
\nocite{Talkin89} \nocite{Nash80} \nocite{Carlson83} \nocite{Cutler83b}
\nocite{Prince84} \nocite{Lehiste80b} \nocite{DHouse90a}
\nocite{Roe91}
\nocite{Cutler77a}
\nocite{Scherer84}
\nocite{Taylor87}
\nocite{Bolinger82}
\nocite{Oakeshott84}
\nocite{Cruz-Ferreira83}
\nocite{Stockwell71}
\nocite{Beckman86a}
\nocite{Beckman86b}
\nocite{Fry58}


\begin{thebibliography}{}

\bibitem[Akmajian and Jackendoff, 1970]{Akmajian70}
Akmajian, A. and Jackendoff, R. (1970).
\newblock Coreferentiality and stress.
\newblock {\em Linguistic Inquiry}, 1(1):124--126.

\bibitem[Allerton and Cruttenden, 1979]{Allerton79}
Allerton, D. and Cruttenden, A. (1979).
\newblock Three reasons for accenting a definite subject.
\newblock {\em Journal of Linguistics}, 15(1):49--53.

\bibitem[Altenberg, 1987]{Altenberg87}
Altenberg, B. (1987).
\newblock {\em Prosodic Patterns in Spoken {E}nglish: Studies in the
  Correlation between Prosody and Grammar for Text-to-Speech Conversion},
  volume~76 of {\em Lund Studies in {E}nglish}.
\newblock Lund University Press, Lund.

\bibitem[Anderson et~al., 1984]{Anderson84}
Anderson, M.~D., Pierrehumbert, J.~B., and Liberman, M.~Y. (1984).
\newblock Synthesis by rule of {E}nglish intonation patterns.
\newblock In {\em Proceedings of the International Conference on Acoustics,
  Speech, and Signal Processing}, volume~1, pages 2.8.1--2.8.4, San Diego.
  ICASSP84.

\bibitem[Avesani and Vayra, 1988]{Avesani88}
Avesani, C. and Vayra, M. (1988).
\newblock Discorso, segmenti di discorso e un{'} ipotesi sull{'} intonazione.
\newblock In {\em Att del Convegno Internazionale {"}Sull'Interpunzione{"}},
  Florence.

\bibitem[Ayers, 1992]{Ayers92}
Ayers, G.~M. (1992).
\newblock Discourse functions of pitch range in spontaneous and read speech.
\newblock Presented at the Linguistic Society of America Annual Meeting.

\bibitem[Baart, 1987]{Baart87}
Baart, J.~L.~G. (1987).
\newblock {\em Focus, Syntax and Accent Placement}.
\newblock PhD thesis, University of Leyden, Leyden.

\bibitem[Bachenko and Fitzpatrick, 1990]{Bachenko90}
Bachenko, J. and Fitzpatrick, E. (1990).
\newblock A computational grammar of discourse-neutral prosodic phrasing in
  {E}nglish.
\newblock {\em Computational Linguistics}, 16(3):155--170.

\bibitem[Ball and Prince, 1977]{Ball77}
Ball, C.~N. and Prince, E.~F. (1977).
\newblock A note on stress and presupposition.
\newblock {\em Linguistic Inquiry}, 8(3):585.

\bibitem[Bardovi-Harlig, 1983a]{Bardovi83a}
Bardovi-Harlig, K. (1983a).
\newblock {\em A Functional Approach to {E}nglish Sentence Stress}.
\newblock PhD thesis, University of Chicago, Chicago IL.

\bibitem[Bardovi-Harlig, 1983b]{Bardovi83b}
Bardovi-Harlig, K. (1983b).
\newblock Pronouns: When `given' and `new' coincide.
\newblock In {\em Papers from the 18th Regional Meeting}. Chicago Linguistic
  Society.

\bibitem[Beach, 1991]{Beach91}
Beach, C. (1991).
\newblock The interpretation of prosodic patterns at points of syntactic
  structure ambiguity: Evidence for cue trading relations.
\newblock {\em Journal of Memory and Language.}, 30:644--663.

\bibitem[Beckman, 1986]{Beckman86c}
Beckman, M. (1986).
\newblock {\em Stress and Non-Stress Accent}.
\newblock Foris Publications.

\bibitem[Beckman and Pierrehumbert, 1986a]{Beckman86b}
Beckman, M. and Pierrehumbert, J. (1986a).
\newblock Intonational structure in {J}apanese and {E}nglish.
\newblock {\em Phonology Yearbook}, 3:15--70.

\bibitem[Beckman and Pierrehumbert, 1986b]{Beckman86a}
Beckman, M. and Pierrehumbert, J. (1986b).
\newblock {J}apanese prosodic phrasing and intonation synthesis.
\newblock In {\em Proceedings of the 24th Annual Meeting}, pages 173--180, New
  York. Association for Computational Linguistics.

\bibitem[Benoit and Bailly, 1989]{Autrans90}
Benoit, C. and Bailly, G., editors (1989).
\newblock {\em Proceedings of the {E}uropean Speech Communication Association
  Workshop on Speech Synthesis}, Autrans. {E}uropean Speech Communication
  Association.

\bibitem[Bing, 1979]{Bing79b}
Bing, J. (1979).
\newblock {\em Aspects of {E}nglish Prosody}.
\newblock PhD thesis, University of Massachusetts at Amherst, Amherst MA.
\newblock Distributed by the Indiana University Linguistics Club.

\bibitem[Bolinger, 1951]{Bolinger51}
Bolinger, D. (1951).
\newblock Intonation: Levels versus configurations.
\newblock {\em Word}, 7(1):199--210.

\bibitem[Bolinger, 1958]{Bolinger58}
Bolinger, D. (1958).
\newblock A theory of pitch accent in {E}nglish.
\newblock {\em Word}, 14:109--149.

\bibitem[Bolinger, 1961]{Bolinger61}
Bolinger, D. (1961).
\newblock Contrastive accent and contrastive stress.
\newblock {\em Language}, 37:83--96.

\bibitem[Bolinger, 1972]{Bolinger72a}
Bolinger, D. (1972).
\newblock Accent is predictable (if you're a mindreader).
\newblock {\em Language}, 48:633--644.

\bibitem[Bolinger, 1982]{Bolinger82}
Bolinger, D. (1982).
\newblock The network tone of voice.
\newblock {\em Journal of Broadcasting}, 26:725--728.

\bibitem[Bolinger, 1986]{Bolinger86}
Bolinger, D. (1986).
\newblock {\em Intonation and Its Parts: Melody in Spoken {E}nglish}.
\newblock Stanford University Press, Palo Alto CA.

\bibitem[Bolinger, 1989]{Bolinger89}
Bolinger, D. (1989).
\newblock {\em Intonation and Its Uses: Melody in Grammar and Discourse}.
\newblock Edward Arnold, London.

\bibitem[Bouton, 1982]{Bouton82}
Bouton, L.~F. (1982).
\newblock Stem polarity and tag intonation in the derivation of the imperative
  tag.
\newblock In Schneider, R., Tuite, K., and Chametzky, R., editors, {\em Papers
  from the Parasession on Nondeclaratives}, pages 23--42, Chicago. Chicago
  Linguistic Society.

\bibitem[Brazil et~al., 1980]{Brazil80}
Brazil, D., Coulthard, M., and Johns, C. (1980).
\newblock {\em Discourse Intonation and Language Teaching}.
\newblock Longman, London.

\bibitem[Bresnan, 1971]{Bresnan71}
Bresnan, J. (1971).
\newblock Sentence stress and syntactic transformations.
\newblock {\em Language}, 47:257--281.

\bibitem[Brown, 1983]{GBrown83a}
Brown, G. (1983).
\newblock Prosodic structure and the given/new distinction.
\newblock In Ladd, D.~R. and Cutler, A., editors, {\em Prosody: Models and
  Measurements}, pages 67--78. Springer Verlag, Berlin.

\bibitem[Brown et~al., 1980]{GBrown80}
Brown, G., Currie, K., and Kenworthy, J. (1980).
\newblock {\em Questions of Intonation}.
\newblock University Park Press, Baltimore.

\bibitem[Brown and Yule, 1983]{GBrown83b}
Brown, G. and Yule, G. (1983).
\newblock {\em Discourse Analysis}.
\newblock Cambridge University Press, Cambridge UK.

\bibitem[Bruce et~al., 1990]{Bruce90}
Bruce, G., Granstrom, B., and House, D. (1990).
\newblock Prosodic phrasing in {S}wedish speech synthesis.
\newblock In {\em Proceedings of the {E}uropean Speech Communication
  Association Workshop on Speech Synthesis}, pages 125--128, Autrans, France.
  {E}uropean Speech Communication Association.

\bibitem[Butterworth, 1975]{Butterworth75}
Butterworth, B. (1975).
\newblock Hesitation and semantic planning in speech.
\newblock {\em Journal of Psycholinguistic Research}, 4:75--87.

\bibitem[Butterworth, 1977]{Butterworth77}
Butterworth, B. (1977).
\newblock Speech and interaction in sound-only communication channels.
\newblock {\em Semiotica}, 20(2):81--99.

\bibitem[Buxton, 1983]{Buxton83}
Buxton, H. (1983).
\newblock Temporal predictability in the perception of {E}nglish speech.
\newblock In Cutler, A. and R, L.~D., editors, {\em Prosody: Models and
  Measurements}, pages 111--122. Springer-Verlag, Berlin.

\bibitem[Cantrall, 1969]{Cantrall69}
Cantrall, W. (1969).
\newblock Pitch, stress, and grammatical relations.
\newblock In {\em Papers from the 5th Regional Meeting}. Chicago Linguistic
  Society.

\bibitem[Cantrall, 1975]{Cantrall75}
Cantrall, W. (1975).
\newblock Favored structures and intonational limitations.
\newblock In {\em Papers from the Parasession on Functionalism}. Chicago
  Linguistic Society.

\bibitem[Carlson, 1983]{Carlson83}
Carlson, L. (1983).
\newblock {\em Dialogue Games}.
\newblock Reidel.

\bibitem[Chafe, 1976]{Chafe76}
Chafe, W. (1976).
\newblock Givenness, contrastiveness, definiteness, subjects, topics, and point
  of view.
\newblock In Li, C., editor, {\em Subject and Topic}, pages 25--55. Academic
  Press, New York.

\bibitem[Cooper and Paccia-Cooper, 1980]{Cooper80}
Cooper, W. and Paccia-Cooper, J. (1980).
\newblock {\em Syntax and Speech}.
\newblock Harvard University Press, Cambridge MA.

\bibitem[Cooper and Sorenson, 1977]{Cooper77}
Cooper, W.~E. and Sorenson, J.~M. (1977).
\newblock Fundamental frequency contours at syntactic boundaries.
\newblock {\em Journal of the Acoustical Society of America}, 62(3):683--692.

\bibitem[Couper-Kuhlen, 1984]{Couper-Kuhlen84}
Couper-Kuhlen, E. (1984).
\newblock A new look at contrastive intonation.
\newblock In Watts, R.~J. and Weidmann, U., editors, {\em Modes of
  Interpretation: Essays Presented to Ernst Leisi}, pages 137--158. Gunter Narr
  Verlag, Tubingen.

\bibitem[Couper-Kuhlen, 1986]{Couper-Kuhlen86}
Couper-Kuhlen, E. (1986).
\newblock {\em An Introduction to {E}nglish Prosody}.
\newblock Edward Arnold, London.

\bibitem[Cruttenden, 1986]{Cruttenden86}
Cruttenden, A. (1986).
\newblock {\em Intonation}.
\newblock Cambridge University Press, Cambridge UK.

\bibitem[Cruz-Ferreira, 1983]{Cruz-Ferreira83}
Cruz-Ferreira (1983).
\newblock {\em Non-Native Comprehension of Intonation Patterns in Portuguese
  and in {E}nglish}.
\newblock PhD thesis, University of Manchester.

\bibitem[Crystal, 1969]{Crystal69}
Crystal, D. (1969).
\newblock {\em Prosodic Systems and Intonation in {E}nglish}.
\newblock Cambridge University Press, Cambridge UK.

\bibitem[Crystal, 1975]{Crystal75}
Crystal, D. (1975).
\newblock {\em The {E}nglish Tone of Voice: Essays in Intonation, Prosody, and
  Paralanguage}.
\newblock Edward Arnold, London.

\bibitem[Culicover and Rochemont, 1983]{Culicover83}
Culicover, P.~W. and Rochemont, M. (1983).
\newblock Stress and focus in {E}nglish.
\newblock {\em Language}, 59(1):123--165.

\bibitem[Cutler, 1977]{Cutler77a}
Cutler, A. (1977).
\newblock The context-dependence of `intonational meanings'.
\newblock In {\em Papers of the Fifth Regional Meeting}, Chicago. Chicago
  Linguistic Society.

\bibitem[Cutler, 1983]{Cutler83b}
Cutler, A. (1983).
\newblock Speakers' conceptions of the functions of prosody.
\newblock In Cutler, A. and Ladd, D.~R., editors, {\em Prosody: Models and
  Measurements}, pages 79--92. Springer-Verlag, Berlin.

\bibitem[Cutler and Foss, 1977]{Cutler77b}
Cutler, A. and Foss, D. (1977).
\newblock On the role of sentence stress in sentence processing.
\newblock {\em Language and Speech}, 20:1--10.

\bibitem[Cutler and Ladd, 1983]{Cutler83a}
Cutler, A. and Ladd, D.~R., e. (1983).
\newblock {\em Prosody: Models and Measurements}.
\newblock Springer-Verlag, Berlin.

\bibitem[Danlos et~al., 1986]{Danlos86}
Danlos, L., La{P}orte, E., and Emerard, F. (1986).
\newblock Synthesis of spoken messages from semantic representations.
\newblock In {\em Proceedings of the 11th International Conference on
  Computational Linguistics}, pages 599--604. International Conference on
  Computational Linguistics.

\bibitem[Davis and Hirschberg, 1988]{Davis88}
Davis, J.~R. and Hirschberg, J. (1988).
\newblock Assigning intonational features in synthesized spoken directions.
\newblock In {\em Proceedings of the 26th Annual Meeting}, pages 187--193,
  Buffalo. Association for Computational Linguistics.

\bibitem[Denes and Pinson, 1973]{Denes73}
Denes, P.~B. and Pinson, E.~N. (1973).
\newblock {\em The Speech Chain: The Physics and Biology of Spoken Language}.
\newblock Anchor Press, Garden City NY.

\bibitem[Dirksen, 1992]{Dirksen92}
Dirksen, A. (1992).
\newblock Accenting and deaccenting: A declarative approach.
\newblock In {\em Proceedings of COLING-92}, pages 865--869.

\bibitem[Downing, 1970]{Downing70}
Downing, B. (1970).
\newblock {\em Syntactic Structure and Phonological Phrasing in {E}nglish}.
\newblock PhD thesis, University of Texas, Austin.

\bibitem[Duncan and Fiske, 1982]{Duncan77}
Duncan, S. and Fiske, D.~W. (1982).
\newblock {\em Face to Face Interaction: Research, Methods and Theory}.
\newblock Lawrence Erlbaum Associates, Hillsdale NJ.

\bibitem[Edwards and Beckman, 1988]{Edwards88}
Edwards, J. and Beckman, M.~E. (1988).
\newblock Articulatory timing and the prosodic interpretation of syllable
  duration.
\newblock {\em Phonetica}, 45:156--174.

\bibitem[Enkvist, 1979]{Enkvist79}
Enkvist, N. (1979).
\newblock Marked focus: Functions and constraints.
\newblock In Greenbaum, S., Leech, G., and Svartvik, J., editors, {\em Studies
  in {E}nglish Linguistics for Randolph Quirk}, pages 134--152. Longmans,
  London.

\bibitem[Erteschik-Shir and Lappin, 1983]{Erteschik83}
Erteschik-Shir, N. and Lappin, S. (1983).
\newblock Under stress: A functional explanation of {E}nglish sentence stress.
\newblock {\em Journal of Linguistics}, 19:419--453.

\bibitem[Ervin-Tripp, 1979]{Ervin-Tripp79}
Ervin-Tripp, S. (1979).
\newblock Children's verbal turn-taking.
\newblock In Ochs, E. and Shieffelin, B.~B., editors, {\em Developmental
  Pragmatics}, pages 391--414. Academic Press, New York.

\bibitem[Fernald, 1991]{Fernald91b}
Fernald, A. (1991).
\newblock Prosody in speech to children: Prelinguistic and linguistic
  functions.
\newblock In Vasta, R., editor, {\em Annals of Child Development}. Jessica
  Kingsley Publishers, London.

\bibitem[Fernald and Mazzie, 1991]{Fernald91a}
Fernald, A. and Mazzie, C. (1991).
\newblock Prosody and focus in speech to infants and adults.
\newblock {\em Developmental Psychology}, 27(2):209--221.

\bibitem[Fowler and Housum, 1987]{Fowler87}
Fowler, C.~A. and Housum, J. (1987).
\newblock Talkers' signaling of {"new"} and {"old"} words in speech and
  listeners' perception and use of the distinction.
\newblock {\em Journal of Memory and Language}, 26:489--504.

\bibitem[Fry, 1958]{Fry58}
Fry, D.~B. (1958).
\newblock Experiments in the perception of stress.
\newblock {\em Language and Speech}, 1:126--152.

\bibitem[Fry, 1979]{Fry79}
Fry, D.~B. (1979).
\newblock {\em The Physics of Speech}.
\newblock Cambridge University Press, Cambridge UK.

\bibitem[Fuchs, 1980]{Fuchs80}
Fuchs, A. (1980).
\newblock Accented subjects in `all-new' utterances.
\newblock In Brettschneider, G. and Lehmann, C., editors, {\em Wege zur
  Universalienforschung: sprachwissenschaftliche Beitrage zum 60.} Narr,
  Tubingen.

\bibitem[Fuchs, 1984]{Fuchs84}
Fuchs, A. (1984).
\newblock Deaccenting and default accent.
\newblock In Gibbon, D. and Richter, H., editors, {\em Intonation, Accent and
  Rhythm}, pages 134--164. Walter de Gruyter, Berlin.

\bibitem[Garding, 1983]{Garding83}
Garding, E. (1983).
\newblock A genertive model of intonation.
\newblock In Cutler, A. and Ladd, D.~R., editors, {\em Prosody: Models and
  Measurements}, pages 11--26. Springer-Verlag.

\bibitem[Gazdar, 1980]{Gazdar80b}
Gazdar, G. (1980).
\newblock Pragmatic constraints on linguistic production.
\newblock In Butterword, B., editor, {\em Language Production}, pages 49--68.
  Academic Press.
\newblock Vol. 1: Speech and Talk.

\bibitem[Gee and Grosjean, 1983]{Gee83}
Gee, J.~P. and Grosjean, F. (1983).
\newblock Performance structure: A psycholinguistic and linguistic apprasial.
\newblock {\em Cognitive Psychology}, 15:411--458.

\bibitem[Gibbon and Richter, 1984]{Gibbon84}
Gibbon, D. and Richter, H., editors (1984).
\newblock {\em Intonation, Accent, and Rhythm: Studies in Discourse Phonology}.
\newblock Walter de Gruyter, Berlin.

\bibitem[Gleitman, 1961]{Gleitman61}
Gleitman, L. (1961).
\newblock Pronominals and stress in {E}nglish.
\newblock {\em Language Learning}, 11:157--169.

\bibitem[Goodwin, 1981]{Goodwin81}
Goodwin, C. (1981).
\newblock {\em Conversational Organization: Interaction between Speakers and
  Hearers}.
\newblock Academic Press, New York.

\bibitem[Grosjean and Gee, 1987]{Grosjean87}
Grosjean, F. and Gee, J.~P. (1987).
\newblock Prosodic structure and spoken work recognition.
\newblock {\em Cognition}, 25:135--155.

\bibitem[Grosjean et~al., 1979]{Grosjean79}
Grosjean, F., Grosjean, L., and Lane, H. (1979).
\newblock The patterns of silence: Performance structures in sentence
  production.
\newblock {\em Cognitive Psychology}, 11:58--81.

\bibitem[Grosz and Hirschberg, 1992]{Grosz92}
Grosz, B. and Hirschberg, J. (1992).
\newblock Some intonational characteristics of discourse structure.
\newblock In {\em Proceedings of the International Conference on Spoken
  Language Processing}, Banff. ICSLP.

\bibitem[Gumperz, 1977]{Gumperz77}
Gumperz, J.~J. (1977).
\newblock Sociocultural knowledge in conversational inference.
\newblock In Saville-Troike, M., editor, {\em Linguistics and Anthropology},
  pages 191--211. Georgetown University Press, Washington.

\bibitem[Gundel, 1978]{Gundel78}
Gundel, J. (1978).
\newblock Stress, pronominalization, and the given-new distinction.
\newblock {\em University of Hawaii Working Papers in Linguistics},
  10(2):1--13.

\bibitem[Gussenhoven, 1983]{Gussenhoven83a}
Gussenhoven, C. (1983).
\newblock {\em On the Grammar and Semantics of Sentence Accents}.
\newblock Foris Publications, Dordrecht.

\bibitem[Halliday, 1967]{Halliday67b}
Halliday, M.~A.~K. (1967).
\newblock {\em Intonation and Grammar in British {E}nglish}.
\newblock Mouton, The Hague.

\bibitem[Halliday and Hassan, 1976]{Halliday76}
Halliday, M.~A.~K. and Hassan, R. (1976).
\newblock {\em Cohesion in {E}nglish}.
\newblock Longman.

\bibitem[Harries-Delisle, 1978]{Harries-Delisle78}
Harries-Delisle, H. (1978).
\newblock Contrastive emphasis and cleft sentences.
\newblock In Greenberg, J., editor, {\em Universals of Human Language}, pages
  419--486. Stanford University Press.

\bibitem[Hart et~al., 1990]{tHart90}
Hart, J., Collier, R., and Cohen, A. (1990).
\newblock {\em A Perceptual Study of Intonation}.
\newblock Cambridge University Press, Cambridge UK.

\bibitem[Hemphill et~al., 1990]{Hemphill90}
Hemphill, C.~T., Godfrey, J.~J., and Doddington, G.~R. (1990).
\newblock The atis spoken language systems pilot corpus.
\newblock In {\em Proceedings of the Speech and Natural Language Workshop},
  pages 96--101, Hidden Valley PA. {DARPA}.

\bibitem[Hirschberg, 1989]{Hirschberg89}
Hirschberg, J. (1989).
\newblock Distinguishing questions by contour in speech recognition tasks.
\newblock In {\em Proceedings of the Speech and Natural Language Workshop}.
  Morgan Kaufmann, Cape Cod MA.

\bibitem[Hirschberg, 1990a]{Hirschberg90}
Hirschberg, J. (1990a).
\newblock Accent and discourse context: Assigning pitch accent in synthetic
  speech.
\newblock In {\em Proceedings of the Eighth National Conference}, pages
  952--957, Boston. American Association for Artificial Intelligence.

\bibitem[Hirschberg, 1990b]{Hirschberg90a}
Hirschberg, J. (1990b).
\newblock Using discourse context to guide pitch accent decisions in synthetic
  speech.
\newblock In {\em Proceedings of the {E}uropean Speech Communication
  Association Workshop on Speech Synthesis}, pages 181--184, Autrans, France.
  {E}uropean Speech Communication Association.

\bibitem[Hirschberg and Grosz, 1992]{Hirschberg92b}
Hirschberg, J. and Grosz, B. (1992).
\newblock Intonational features of local and global discourse structure.
\newblock In {\em Proceedings of the Speech and Natural Language Workshop},
  pages 441--446, Harriman NY. DARPA, Morgan Kaufmann.

\bibitem[Hirschberg and Litman, 1987]{Hirschberg87a}
Hirschberg, J. and Litman, D. (1987).
\newblock Now let's talk about {\em now}: Identifying cue phrases
  intonationally.
\newblock In {\em Proceedings of the 25th Annual Meeting}, pages 163--171,
  Stanford University. Association for Computational Linguistics.

\bibitem[Hirschberg and Pierrehumbert, 1986]{Hirschberg86}
Hirschberg, J. and Pierrehumbert, J. (1986).
\newblock The intonational structuring of discourse.
\newblock In {\em Proceedings of the 24th Annual Meeting}, pages 136--144, New
  York. Association for Computational Linguistics.

\bibitem[Hirschberg and Ward, 1991a]{Hirschberg91a}
Hirschberg, J. and Ward, G. (1991a).
\newblock Accent and bound anaphora.
\newblock {\em Cognitive Linguistics}, 2(2):101--121.

\bibitem[Hirschberg and Ward, 1991b]{Hirschberg91b}
Hirschberg, J. and Ward, G. (1991b).
\newblock The influence of pitch range, duration, amplitude, and spectral
  features on the interpretation of {\bf l*+h l h\%}.
\newblock {\em Journal of Phonetics}.

\bibitem[Hirsh-Pasek et~al., 1985]{Hirsh-Pasek85}
Hirsh-Pasek, K., Nelson, D.~K., Jusczyk, P., Wright, K., and Druss, B. (1985).
\newblock Clauses are perceptual units for prelinguistic infants.
\newblock Paper presented at the Tenth Annual Boston University Conference on
  Language Development.

\bibitem[Hirst, 1983]{Hirst83}
Hirst, D. (1983).
\newblock Structures and categories in prosodic representations.
\newblock In Cutler, A. and Ladd, D.~R., editors, {\em Prosody: Models and
  Measurements}, pages 93--110. Springer-Verlag, Berlin.

\bibitem[Horne, 1985]{Horne85}
Horne, M. (1985).
\newblock {E}nglish sentence stress, grammatical functions and contextual
  coreference.
\newblock {\em Studia Linguistica}, 39:51--66.

\bibitem[Horne, 1987]{Horne87}
Horne, M. (1987).
\newblock Towards a discourse-based model of {E}nglish sentence intonation.
\newblock Working Papers~32, Lund University Department of Linguistics.

\bibitem[Horne, 1991a]{Horne91a}
Horne, M. (1991a).
\newblock Accentual patterning in `new' vs `given' subjects in {E}nglish.
\newblock Working Papers~36, Department of Linguistics, Lund University, Lund.

\bibitem[Horne, 1991b]{Horne91b}
Horne, M. (1991b).
\newblock Phonetic correlates of the new/given parameter.
\newblock In {\em Proceedings of the Twelfth International Congress of Phonetic
  Sciences}, pages 230--233, Aix-en-Provence. ICPhS.

\bibitem[House, 1989]{DHouse89}
House, D. (1989).
\newblock Automatic recognition of prosodic categories.
\newblock In Szende, T., editor, {\em Proceedings of the Speech Research '89
  International Conference}, pages 347--350, Budapest.

\bibitem[House, 1990a]{DHouse90a}
House, D. (1990a).
\newblock {\em Tonal Perception in Speech}, volume~24 of {\em Travaux de
  l'Institut de Linguistique de Lund}.
\newblock Lund University Press, Lund.

\bibitem[House, 1990b]{JHouse90a}
House, J. (1990b).
\newblock A revised model for intonation for synthesis by rule.
\newblock Speech, Hearing and Language: Work in Progress U.~C.~L. No. 4,
  University College London, London.

\bibitem[House and Youd, 1990]{JHouse90b}
House, J. and Youd, N. (1990).
\newblock Contextually appropriate intonation in speech synthesis.
\newblock In {\em Proceedings of the {E}uropean Speech Communication
  Association Workshop on Speech Synthesis}, pages 185--188, Autrans.
  {E}uropean Speech Communication Association.

\bibitem[Jackendoff, 1972]{Jackendoff72}
Jackendoff, R.~S. (1972).
\newblock {\em Semantic Interpretation in Generative Grammar}.
\newblock MIT Press, Cambridge MA.

\bibitem[Klatt, 1975]{Klatt75a}
Klatt, D. (1975).
\newblock Vowel lengthening is syntactically determined in connected discourse.
\newblock {\em Journal of Phonetics}, 3:129--140.

\bibitem[Klatt, 1987]{Klatt87}
Klatt, D.~H. (1987).
\newblock Review of text-to-speech conversion for {E}nglish.
\newblock {\em Journal of the Acoustical Society of America}, 82(3):737--793.

\bibitem[Kruyt, 1985]{Kruyt85}
Kruyt, J.~G. (1985).
\newblock {\em Accents from Speakers to Listeners: An Experimental Study of the
  Production and Perception of Accent Patterns in {D}utch}.
\newblock PhD thesis, University of Leyden.

\bibitem[Kumpf, 1987]{Kumpf87}
Kumpf, L.~E. (1987).
\newblock The use of pitch phenomena in the structuring of stories.
\newblock In Tomlin, R.~S., editor, {\em Coherence and Grounding in Discourse},
  volume~11. John Benjamins Publishing Company.

\bibitem[Kutik et~al., 1983]{Kutik83}
Kutik, E.~J., Cooper, W.~E., and Boyce, S. (1983).
\newblock Declination of fundamental frequency in speaker's production of
  parenthetical and main clauses.
\newblock {\em Journal of the Acoustical Society of America}, 73:1731--1738.

\bibitem[Ladd, 1977]{Ladd77}
Ladd, D.~R. (1977).
\newblock The function of the a-rise accent in {E}nglish.
\newblock Distributed by the Indiana University Linguistics Club.

\bibitem[Ladd, 1978]{Ladd78}
Ladd, D.~R. (1978).
\newblock Stylized intonation.
\newblock {\em Language}, 54:517--540.

\bibitem[Ladd, 1980]{Ladd80}
Ladd, D.~R. (1980).
\newblock {\em The Structure of Intonational Meaning}.
\newblock Indiana University Press, Bloomington, Ind.

\bibitem[Ladd, 1983a]{Ladd83c}
Ladd, D.~R. (1983a).
\newblock Levels-vs.-configurations, revisited.
\newblock In et~al, A., editor, {\em Essays in Honor of Charles F. Hockett}.
  E.~J.~Brill, Leiden.

\bibitem[Ladd, 1983b]{Ladd83a}
Ladd, D.~R. (1983b).
\newblock Peak features and overall slope.
\newblock In Cutler, A. and Ladd, D.~R., editors, {\em Prosody: Models and
  Measurements}, pages 39--52. Springer-Verlag.

\bibitem[Ladd, 1983c]{Ladd83b}
Ladd, D.~R. (1983c).
\newblock Phonological features of intonational peaks.
\newblock {\em Language}, 4:721--759.

\bibitem[Ladd, 1986]{Ladd86}
Ladd, D.~R. (1986).
\newblock Intonational phrasing: The case for recursive prosodic structure.
\newblock {\em Phonology Yearbook}, 3:311--340.

\bibitem[Ladd, 1988]{Ladd88}
Ladd, D.~R. (1988).
\newblock Declination `reset' and the hierarchical organization of utterances.
\newblock {\em Journal of the Acoustical Society of America}, 84:530--544.

\bibitem[Ladd and Monaghan, 1987]{Ladd87a}
Ladd, D.~R. and Monaghan, A. (1987).
\newblock Modelling rhythmic and syntactic effects on accent in long noun
  phrases.
\newblock In Laver, J. and Jack, M., editors, {\em Proceedings of the
  {E}uropean Conference on Speech Technology}, pages 29--32, Edinburgh. {CEP}.

\bibitem[Ladd et~al., 1985]{Ladd85}
Ladd, D.~R., Silverman, K.~E.~A., Tolkmitt, F., Bergmann, G., and Scherer,
  K.~R. (1985).
\newblock Evidence for the independent function of intonation contour type,
  voice quality, and f0 range in signaling speaker affect.
\newblock {\em Journal of the Acoustical Society of America}, 78(1):435--444.

\bibitem[Ladefoged, 1962]{Ladefoged62}
Ladefoged, P. (1962).
\newblock {\em Elements of Acoustic Phonetics}.
\newblock The University of Chicago Press, Chicago.

\bibitem[Lakoff, 1971]{Lakoff71}
Lakoff, G. (1971).
\newblock Presupposition and relative well-formedness.
\newblock In {\em Semantics: An Interdisciplinary Reader in Philosophy,
  Linguistics, and Psychology}, pages 329--340. Cambridge University Press,
  Cambridge UK.

\bibitem[Lea, 1979]{Lea79}
Lea, W.~A. (1979).
\newblock Prosodic aids to speech recognition.
\newblock In Lea, W.~A., editor, {\em Trends in Speech Recognition}, pages
  166--205. Prentice-Hall, Englewood Cliffs NJ.

\bibitem[Lea et~al., 1975]{Lea75}
Lea, W.~A., Medress, M.~F., and Skinner, T.~E. (1975).
\newblock A prosodically guided speech understanding system.
\newblock {\em IEEE Trans. Acoust. Speech, Signal Processing}, ASSP-23:30--38.

\bibitem[Lehiste, 1973]{Lehiste73}
Lehiste, I. (1973).
\newblock Phonetic disambiguation of syntactic ambiguity.
\newblock {\em Glossa}, 7:197--222.

\bibitem[Lehiste, 1980a]{Lehiste80b}
Lehiste, I. (1980a).
\newblock Phonetic characteristics of discourse.
\newblock Paper presented at the Meeting of the Committee on Speech Research,
  Acoustical Society of {J}apan.

\bibitem[Lehiste, 1980b]{Lehiste80a}
Lehiste, I. (1980b).
\newblock Phonetic manifestation of syntactic structure in {E}nglish.
\newblock {\em Annual Bulletin of the Research Institute of Logopedics and
  Phoniatrics, University of Tokyo}, 14(1-27).

\bibitem[Lehiste et~al., 1976]{Lehiste76}
Lehiste, I., Olive, J., and Streeter, L. (1976).
\newblock Role of duration in disambiguating syntactically ambiguous sentences.
\newblock {\em Journal of the Acoustical Society of America}, 60:1199--1202.

\bibitem[Lehman, 1977]{Lehman77}
Lehman, C. (1977).
\newblock A re-analysis of givenness: Stress in discourse.
\newblock In {\em Papers from the 13th Annual Meeting}, pages 316--. Chicago
  Linguistic Society.

\bibitem[Liberman and Pierrehumbert, 1984]{Liberman84}
Liberman, M. and Pierrehumbert, J. (1984).
\newblock Intonational invariants under changes in pitch range and length.
\newblock In Aronoff, M. and Oehrle, R., editors, {\em Language Sound
  Structure}. MIT Press, Cambridge.

\bibitem[Liberman and Prince, 1977]{Liberman77}
Liberman, M. and Prince, A. (1977).
\newblock On stress and linguistic rhythm.
\newblock {\em Linguistic Inquiry}, 8(2):249--336.

\bibitem[Liberman and Sag, 1974]{Liberman74}
Liberman, M. and Sag, I.~A. (1974).
\newblock Prosodic form and discourse function.
\newblock In {\em Papers of the Tenth Regional Meeting}, pages 416--427.
  Chicago Linguistic Society.

\bibitem[Liberman, 1975]{Liberman75}
Liberman, M.~Y. (1975).
\newblock {\em The Intonation System of {E}nglish}.
\newblock PhD thesis, MIT, Cambridge MA.

\bibitem[Litman and Hirschberg, 1990]{Litman90}
Litman, D. and Hirschberg, J. (1990).
\newblock Disambiguating cue phrases in text and speech.
\newblock In {\em Papers Presented to the 13th International Conference on
  Computational Linguistics}, pages 251--256, Helsinki. International
  Conference on Computational Linguistics.

\bibitem[Ljolje, 1986]{Ljolje86}
Ljolje, A. (1986).
\newblock {\em Intonation and Phonetic Segmentation Using Hidden Markov
  Models}.
\newblock PhD thesis, University of Cambridge.

\bibitem[Ljolje and Fallside, 1987]{Ljolje87a}
Ljolje, A. and Fallside, F. (1987).
\newblock Modelling of speech using primarily prosodic parameters.
\newblock {\em Computer Speech and Language}, 2:185--204.

\bibitem[Luj\'{a}n, 1985]{Lujan85}
Luj\'{a}n, M. (1985).
\newblock Stress and the binding of pronouns.
\newblock In {\em Papers of the Twenty-first Regional Meeting}, pages 248--262.
  Chicago Linguistic Society.

\bibitem[Marcus and Hindle, 1985]{Marcus85}
Marcus, M.~P. and Hindle, D. (1985).
\newblock A computational account of extra categorial elements in {J}apanese.
\newblock In Kuroda, S.~Y., editor, {\em Papers presented at the First SDF
  Workshop in {J}apanese Syntax}, La Jolla. System Development Foundation.

\bibitem[Marcus and Hindle, 1990]{Marcus90}
Marcus, M.~P. and Hindle, D. (1990).
\newblock Description theory and intonation boundaries.
\newblock In Altmann, G., editor, {\em Computational and Cognitive Models of
  Speech}. MIT Press, Cambridge MA.

\bibitem[Martin, 1970]{Martin70}
Martin, E. (1970).
\newblock Toward an analysis of subjective phrase structure.
\newblock {\em Psychological Bulletin}, 74:153--166.

\bibitem[Misono and Kiritani, 1990]{Misono90}
Misono, Y. and Kiritani, S. (1990).
\newblock The distribution pattern of pauses in lecture-style speech.
\newblock {\em Annual Bulletin of the RILP}, 24:101--111.

\bibitem[Morgan et~al., 1987]{Morgan87}
Morgan, J.~L., Meier, R.~P., and Newport, E.~L. (1987).
\newblock Structural packaging in the input to language learning: Contributions
  of prosodic and morphological marking of phrases to the acquisition of
  language.
\newblock {\em Cognitive Psychology}, 19:498--550.

\bibitem[Nash and Mulac, 1980]{Nash80}
Nash, R. and Mulac, A. (1980).
\newblock The intonation of verifiability.
\newblock In Waugh, L.~R. and van Schooneveld, C.~H., editors, {\em The Melody
  of Language}, pages 219--242. University Park Press, Baltimore.

\bibitem[Nooteboom and Terken, 1982]{Nooteboom82}
Nooteboom, S.~G. and Terken, J. (1982).
\newblock What makes speakers omit pitch accents?: An experiment.
\newblock {\em Phonetica}, 39:317--336.

\bibitem[Oakeshott-Taylor, 1984]{Oakeshott84}
Oakeshott-Taylor, J. (1984).
\newblock Factuality and intonation.
\newblock {\em Journal of Linguistics}, 20:1--21.

\bibitem[O'Connor and Arnold, 1961]{O'Connor61}
O'Connor, J. and Arnold, G. (1961).
\newblock {\em Intonation of Colloquial {E}nglish}.
\newblock Longman, London.

\bibitem[Olive and Liberman, 1985]{Olive85}
Olive, J.~P. and Liberman, M.~Y. (1985).
\newblock Text to speech -- an overview.
\newblock {\em Journal of the Acoustic Society of America, Suppl. 1},
  78(Fall):s6.

\bibitem[O'Malley et~al., 1973]{O'Malley73}
O'Malley, M.~M., Kloker, D., and Dara-Abrams, B. (1973).
\newblock Recovering parentheses from spoken algebraic expressions.
\newblock {\em IEEE Trans. Audio Electroacoust.}, AU-21:217--220.

\bibitem[O'Shaughnessy, 1989]{O'Shaughnessy89}
O'Shaughnessy, D. (1989).
\newblock Parsing with a small dictionary for applictions such as text to
  speech.
\newblock {\em Computational Linguistics}, 15(2):97--108.

\bibitem[Ostendorf et~al., 1990]{Ostendorf90}
Ostendorf, M., Price, P., Bear, J., and Wightman, C.~W. (1990).
\newblock The use of relative duration in syntactic disambiguation.
\newblock In {\em Proceedings of the Speech and Natural Language Workshop},
  pages 26--31, Hidden Valley PA. {DARPA}, Morgan Kaufmann.

\bibitem[Pierrehumbert, 1981]{Pierrehumbert81}
Pierrehumbert, J. (1981).
\newblock Synthesising intonation.
\newblock {\em Journal of the Acoustical Society of America}, 70(4):985--995.

\bibitem[Pierrehumbert, 1989]{Pierrehumbert89}
Pierrehumbert, J. (1989).
\newblock Categories of tonal alignment in {E}nglish.
\newblock {\em Phonetica}.

\bibitem[Pierrehumbert and Hirschberg, 1990]{Pierrehumbert90}
Pierrehumbert, J. and Hirschberg, J. (1990).
\newblock The meaning of intonational contours in the interpretation of
  discourse.
\newblock In Cohen, P., Morgan, J., and Pollack, M., editors, {\em Intentions
  in Communication}. MIT Press, Cambridge MA.

\bibitem[Pierrehumbert, 1980]{Pierrehumbert80}
Pierrehumbert, J.~B. (1980).
\newblock {\em The Phonology and Phonetics of {E}nglish Intonation}.
\newblock PhD thesis, Massachusetts Institute of Technology.
\newblock Distributed by the Indiana University Linguistics Club.

\bibitem[Pierrehumbert, 1983]{Pierrehumbert83}
Pierrehumbert, J.~B. (1983).
\newblock Automatic recognition of intonation patterns.
\newblock In {\em Proceedings of the 21st Annual Meeting}, pages 85--90,
  Cambridge MA. Association for Computational Linguistics.

\bibitem[Pierrehumbert and Steele, 1987]{Pierrehumbert87b}
Pierrehumbert, J.~B. and Steele, S. (1987).
\newblock How many rise-fall-rise contours?
\newblock In {\em Proceedings of the Eleventh Meeting}, Tallinn. International
  Congress of Phonetic Sciences.

\bibitem[Pike, 1945]{Pike45}
Pike, K. (1945).
\newblock {\em The Intonation of {American} {{E}nglish}}.
\newblock University of Michigan Press, Ann Arbor MI.

\bibitem[Price et~al., 1988]{Price88}
Price, P., Fisher, W.~M., Bernstein, J., and Pallett, D.~S. (1988).
\newblock The {{DARPA}} 1000-word {R}esource {M}anagement {D}atabase for
  continuous speech recognition.
\newblock In {\em Proceedings}, volume~1, pages 651--654, New York. ICASSP88.

\bibitem[Prince, 1984]{Prince84}
Prince, E.~F. (1984).
\newblock Language and the law: Reference, stress, and context.
\newblock In Schiffrin, D., editor, {\em GURT84: Meaning, Form and Use in
  Context: Linguistic Applications}, pages 240--252. Georgetown University
  Press, Washington DC.

\bibitem[Quirk et~al., 1964]{Quirk64}
Quirk, R., Svartvik, J., Duckworth, A.~P., Rusiecki, J. P.~L., and Colin, A.
  J.~T. (1964).
\newblock Studies in the correspondence of prosodic to grammatical features in
  {E}nglish.
\newblock In {\em Proceedings of the Ninth International Congress}, pages
  679--691. International Congress of Linguists.

\bibitem[Ratner, 1986]{Ratner86}
Ratner, N.~B. (1986).
\newblock Durational cues which mark clause boundaries in mother-child speech.
\newblock {\em Journal of Phonetics}, 14:303--309.

\bibitem[Rees and Urquhart, 1976]{Rees76}
Rees, M. and Urquhart, A.~H. (1976).
\newblock Intonation as a guide to readers' structuring of prose texts.
\newblock Work in Progress~9, University of Edinburgh, Department of
  Linguistics.

\bibitem[Rochemont and Culicover, 1990]{Rochemont90}
Rochemont, M.~S. and Culicover, P.~W. (1990).
\newblock {\em {E}nglish Focus Constructions and the Theory of Grammar}.
\newblock Cambridge University Press, Cambridge UK.

\bibitem[Roe et~al., 1991]{Roe91}
Roe, D.~B., Pereira, F., Sproat, R.~W., Riley, M.~D., Moreno, P.~J., and
  Macarron, A. (1991).
\newblock Toward a spoken language translator for restricted-domain
  context-free languages.
\newblock In {\em Proceedings of the Second {E}uropean Conference on Speech
  Communication and Technology}, pages 1063--1066, Genova. Eurospeech-91.

\bibitem[Rooth, 1985]{Rooth85}
Rooth, M. (1985).
\newblock {\em Association with Focus}.
\newblock PhD thesis, University of Massachusetts, Amherst MA.

\bibitem[Rooth, 1991]{Rooth91}
Rooth, M. (1991).
\newblock A theory of focus interpretation.
\newblock Presented at the Workshop on the Syntax and Semantics of Focus, Third
  {E}uropean Summer School in Language, Logic and Information, Universitaet des
  Saarlandes, Saarbrucken.

\bibitem[Sag and Liberman, 1975]{Sag75}
Sag, I.~A. and Liberman, M.~Y. (1975).
\newblock The intonational disambiguation of indirect speech acts.
\newblock In {\em Papers from the Eleventh Regional Meeting}. Chicago
  Linguistic Society.

\bibitem[Schegloff, 1979]{Schegloff79b}
Schegloff, E.~A. (1979).
\newblock Identification and recognition in telephone conversation openings.
\newblock In Psathas, G., editor, {\em Everyday Language: Studies in
  Ethnomethodology}, pages 23--78. Irvington, New York.

\bibitem[Scherer et~al., 1984]{Scherer84}
Scherer, K., Ladd, D.~R., and Silverman, K. E.~A. (1984).
\newblock Vocal cues to speaker affect: Testing two models.
\newblock {\em Journal of the Acoustical Society of America}, 76(5):1346--1356.

\bibitem[Schmerling, 1971]{Schmerling71}
Schmerling, S. (1971).
\newblock Presupposition and the notion of normal stress.
\newblock In {\em Papers from the 7th Regional Meeting}, pages 242--253.
  Chicago Linguistic Society.

\bibitem[Schmerling, 1974]{Schmerling74}
Schmerling, S. (1974).
\newblock A re-examination of the notion {NORMAL STRESS}.
\newblock {\em Language}, 50:66--73.

\bibitem[Schmerling, 1975]{Schmerling75b}
Schmerling, S. (1975).
\newblock Evidence from sentence stress for the notions of topic and comment.
\newblock In Schmerling, S. and King, R., editors, {\em Texas Linguistics
  Forum}, pages 135--141. University of Texas, Department of Linguistics.

\bibitem[Schmerling, 1976]{Schmerling76}
Schmerling, S.~F. (1976).
\newblock {\em Aspects of {E}nglish Sentence Stress}.
\newblock University of Texas Press, Austin.
\newblock Revised 1973 thesis, University of Illinois at Urbana.

\bibitem[Selkirk, 1984]{Selkirk84a}
Selkirk, E. (1984).
\newblock {\em Phonology and Syntax}.
\newblock MIT Press, Cambridge MA.

\bibitem[Silva-Corvalan, 1983]{Silva-Corvalan83}
Silva-Corvalan, C. (1983).
\newblock On the interaction of word order and intonation: Some {OV}
  constructions in spanish.
\newblock In Klein, F., editor, {\em Discourse Perspectives on Syntax}, pages
  117--140. Academic Press.

\bibitem[Silverman, 1986]{Silverman86}
Silverman, K. (1986).
\newblock Synthesis and perception of paragraph prosody.
\newblock In {\em Proceedings}. First Australian Conference of Speech Science
  and Technology.

\bibitem[Silverman, 1987]{Silverman87}
Silverman, K. (1987).
\newblock {\em The Structure and Processing of Fundamental Frequency Contours}.
\newblock PhD thesis, Cambridge University, Cambridge UK.

\bibitem[Silverman et~al., 1992]{Silverman92a}
Silverman, K. E.~A., Blaauw, E., Spitz, J., and Pitrelli, J.~F. (1992).
\newblock Towards using prosody in speech recognition/understanding systems:
  Differences between read and spontaneous speech.
\newblock In {\em Proceedings of the Speech and Natural Language Workshop},
  pages 435--440, Harriman NY. DARPA, Morgan Kaufmann.

\bibitem[Sproat, 1990]{Sproat90}
Sproat, R. (1990).
\newblock Stress assignment in complex nominals for {E}nglish text-to-speech.
\newblock In {\em Proceedings of the {E}uropean Speech Communication
  Association Workshop on Speech Synthesis}, pages 129--132, Autrans, France.
  {E}uropean Speech Communication Association.

\bibitem[Steedman, 1990]{Steedman90}
Steedman, M. (1990).
\newblock Structure and intonation in spoken language understanding.
\newblock In {\em Proceedings of the 28th Annual Meeting}, pages 9--16,
  Pittsburgh. Association for Computational Linguistics.

\bibitem[Stockwell, 1971]{Stockwell71}
Stockwell, R.~P. (1971).
\newblock The role of intonation: Reconsiderations and other considerations.
\newblock {\em UCLA Working Papers in Phonetics}, 21:25--49.
\newblock Cited in \cite{Schmerling76}.

\bibitem[Streeter, 1978]{Streeter78}
Streeter, L. (1978).
\newblock Acoustic determinants of phrase boundary perception.
\newblock {\em Journal of the Acoustical Society of America}, 63(6):1582--1592.

\bibitem[Sugito et~al., 1990]{Sugito90}
Sugito, M., Ohyama, G., and Hirose, H. (1990).
\newblock A preliminary study on pauses and breaths in reading speech
  materials.
\newblock {\em Annual Bulletin of the RILP}, 24:121--130.

\bibitem[Taft, 1984]{Taft84}
Taft, L. (1984).
\newblock {\em Prosodic Constraints and Lexical Parsing Strategies}.
\newblock PhD thesis, University of Massachusetts, Amherst MA.

\bibitem[Talkin, 1989]{Talkin89}
Talkin, D. (1989).
\newblock Looking at speech.
\newblock {\em Speech Technology}, 4:74--77.

\bibitem[Taylor and Wales, 1987]{Taylor87}
Taylor, S. and Wales, R. (1987).
\newblock Primitive mechanisms of accent perception.
\newblock {\em Journal of Phonetics}, 15:235--246.

\bibitem[Terken, 1984]{Terken84}
Terken, J. (1984).
\newblock The distribution of pitch accents in instructions as a function of
  discourse structure.
\newblock {\em Language and Speech}, 27:269--289.

\bibitem[Terken and Nooteboom, 1987]{Terken87}
Terken, J. and Nooteboom, S.~G. (1987).
\newblock Opposite effects of accentuation and deaccentuation on verification
  latencies for given and new information.
\newblock {\em Language and Cognitive Processes}, 2(3/4):145--163.

\bibitem[Terken, 1985]{Terken85}
Terken, J.~M.~B. (1985).
\newblock {\em Use and Function of Accentuation: Some Experiments}.
\newblock PhD thesis, University of Leiden, Helmond, Neth.

\bibitem[Umeda, 1982]{Umeda82}
Umeda, N. (1982).
\newblock Boundary: Perceptual and acoustic properties and syntactic and
  statistical determinants.
\newblock {\em Speech and Language}, 7:333--371.

\bibitem[Vaissier, 1983]{Vaissiere83}
Vaissier, J. (1983).
\newblock Language-independent prosodic features.
\newblock In Cutler, A. and Ladd, D.~R., editors, {\em Prosody: Models and
  Measurements}, pages 53--66. Springer-Verlag, Berlin.

\bibitem[Waibel, 1979]{Waibel79}
Waibel, A. (1979).
\newblock The pitch contour in wh-questions.
\newblock {\em Speech Communication Papers}, pages 187--190.
\newblock Acoustical Society of America.

\bibitem[Waibel, 1988]{Waibel88}
Waibel, A. (1988).
\newblock {\em Prosody and Speech Recognition}.
\newblock Pitman Publishing, London.

\bibitem[Waibel and Lee, 1990]{Waibel90a}
Waibel, A. and Lee, K.-F., editors (1990).
\newblock {\em Readings in Speech Recognition}.
\newblock Morgan Kaufmann Publishers, Inc., San Mateo CA.

\bibitem[Wales and Toner, 1979]{Wales79}
Wales, R. and Toner, H. (1979).
\newblock Intonation and ambiguity.
\newblock In Cooper, W.~E. and Walker, E.~C., editors, {\em Sentence
  Processing: Psycholinguistic Studies Presented to Merrill Garrett}. Halsted
  Press, New York.

\bibitem[Wang and Hirschberg, 1991]{Wang91a}
Wang, M.~Q. and Hirschberg, J. (1991).
\newblock Predicting intonational boundaries automatically from text: The
  {ATIS} domain.
\newblock In {\em Proceedings of the Speech and Natural Language Workshop},
  pages 378--383, Pacific Grove CA. {DARPA}, Morgan Kaufmann.

\bibitem[Wang and Hirschberg, 1992]{Wang92}
Wang, M.~Q. and Hirschberg, J. (1992).
\newblock Automatic classification of intonational phrase boundaries.
\newblock {\em Computer Speech and Language}, 6:175--196.

\bibitem[Ward and Hirschberg, 1985]{Ward85b}
Ward, G. and Hirschberg, J. (1985).
\newblock Implicating uncertainty: The pragmatics of fall-rise intonation.
\newblock {\em Language}, 61(4):747--776.

\bibitem[Wilkenfeld, 1981]{Wilkenfeld81}
Wilkenfeld, D. (1981).
\newblock Reading, prosody, and orthography.
\newblock Status Report on Speech Research SR-67/68, Haskins Laboratories.

\bibitem[Williams, 1980]{Williams80}
Williams, E. (1980).
\newblock Remarks on stress and anaphora.
\newblock {\em Journal of Linguistic Research}, 1(3).

\bibitem[Wilson and Sperber, 1979]{Wilson79}
Wilson, D. and Sperber, D. (1979).
\newblock Ordered entailments: An alternative to presuppositional theories.
\newblock In Oh, C.-K. and Dinneen, D.~A., editors, {\em Syntax and Semantics},
  volume~11, pages 229--324. Academic Press, New York.

\bibitem[Witten and Madams, 1977]{LWitten77}
Witten, L. and Madams, P. (1977).
\newblock The telephone inquiry service: A man-machine system using synthetic
  speech.
\newblock {\em International Journal of Man-Machine Studies}, 9:449--464.

\bibitem[Young and Fallside, 1979]{Young79}
Young, S.~J. and Fallside, F. (1979).
\newblock Speech synthesis from concept: A method for speech output from
  information systems.
\newblock {\em Journal of the Acoustic Society of America}, 66(3):685--695.

\bibitem[Yule, 1980]{Yule80}
Yule, G. (1980).
\newblock Speakers' topics and major paratones.
\newblock {\em Lingua}, 52:33--47.

\bibitem[Zacharski, 1992]{Zacharski92}
Zacharski, R. (1992).
\newblock Generation of accent in nominally premodified noun phrases.
\newblock In {\em Papers Presented to the 15th International Conference on
  Computational Linguistics}, pages 253--259, Nantes. International Conference
  on Computational Linguistics.

\end{thebibliography}

\end{document}